\documentclass{article}

\usepackage[preprint]{corl_2024} 
\usepackage{times}

\usepackage[numbers]{natbib}
\usepackage{multicol}
\usepackage{wrapfig}
\usepackage{amsmath}
\usepackage{amssymb}
\usepackage{mathtools}
\usepackage{microtype}
\usepackage{graphicx}
\usepackage{subcaption}
\usepackage{booktabs} 
\usepackage{dsfont}

\usepackage{times}

\usepackage{xcolor} 
\usepackage{bm}
\usepackage{xspace}
\usepackage{caption}
\usepackage{multirow}
\usepackage{url} 
\hypersetup{
breaklinks=true,colorlinks,citecolor=blue
}

\usepackage{subcaption}
\usepackage{float}
\usepackage[title]{appendix}
\newcommand{\figref}[1]{Figure~\ref{#1}}
\newcommand{\tblref}[1]{Table~\ref{#1}}
\newcommand{\secref}[1]{Section~\ref{#1}}
\newcommand{\stopsign}{I^{stop}_t}
\newcommand{\expsymbol}{\mathcal{E}}
\newcommand{\stusymbol}{\mathcal{S}}
\newcommand{\expert}{\pi^\expsymbol}
\newcommand{\obexpert}{\bm{o}^\expsymbol}
\newcommand{\student}{\pi^\stusymbol}
\newcommand{\obstudent}{\bm{o}^\stusymbol}
\newlength{\charwidth}
\newcommand{\legalTM}{\textsuperscript{\texttrademark}}

\newcommand{\videourl}{\url{https://taochenshh.github.io/projects/veg-peeling}}

\title{Vegetable Peeling: A Case Study in Constrained Dexterous Manipulation}

\author{
  Tao Chen$^1$, Eric Cousineau$^2$, Naveen Kuppuswamy$^2$, Pulkit Agrawal$^1$\\
  $^1$Massachusetts Institute of Technology, $^2$Toyota Research Institute\\
 \tt{\{taochen, pulkitag\}@mit.edu}
}

\begin{document}
\maketitle

\vspace{-1.2cm}
\begin{figure}[!h]
    \centering
    \includegraphics[width=\linewidth]{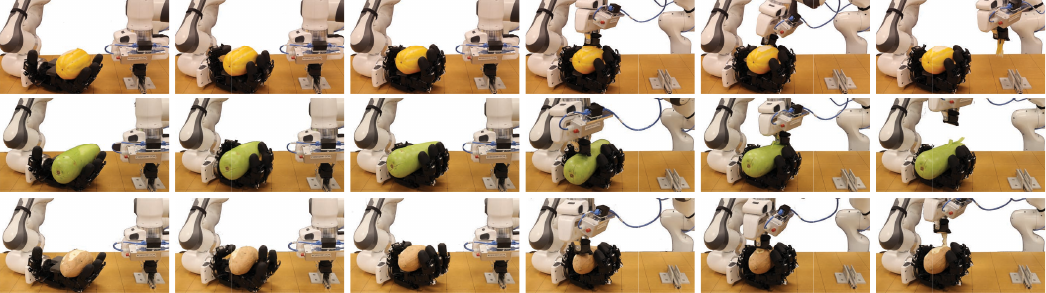}
    \caption{We present a dexterous manipulation system that utilizes an Allegro hand mounted on a Franka robot arm to reorient food items for downstream peeling. The other Franka robot arm (the right arm in the figure) uses its gripper to grasp a peeler for peeling. The reorientation controller for the Allegro hand is learned through reinforcement learning, while the peeling is performed via teleoperation. In the figure, we demonstrate the process of reorienting and peeling a melon, a sweet potato, and a squash from top to bottom row.}
    \label{fig:teaser}
\end{figure}
\vspace{-0.5cm}
\begin{abstract}
Recent studies have made significant progress in addressing dexterous manipulation problems, particularly in in-hand object reorientation. However, there are few existing works that explore the potential utilization of developed dexterous manipulation controllers for downstream tasks. In this study, we focus on constrained dexterous manipulation for food peeling. Food peeling presents various constraints on the reorientation controller, such as the requirement for the hand to securely hold the object after reorientation for peeling. We propose a simple system for learning a reorientation controller that facilitates the subsequent peeling task. Videos are available at: \videourl.
\end{abstract}


\section{Introduction}
\begin{wrapfigure}{r}{0.3\textwidth}
  \begin{center}
    \includegraphics[width=0.99\linewidth]{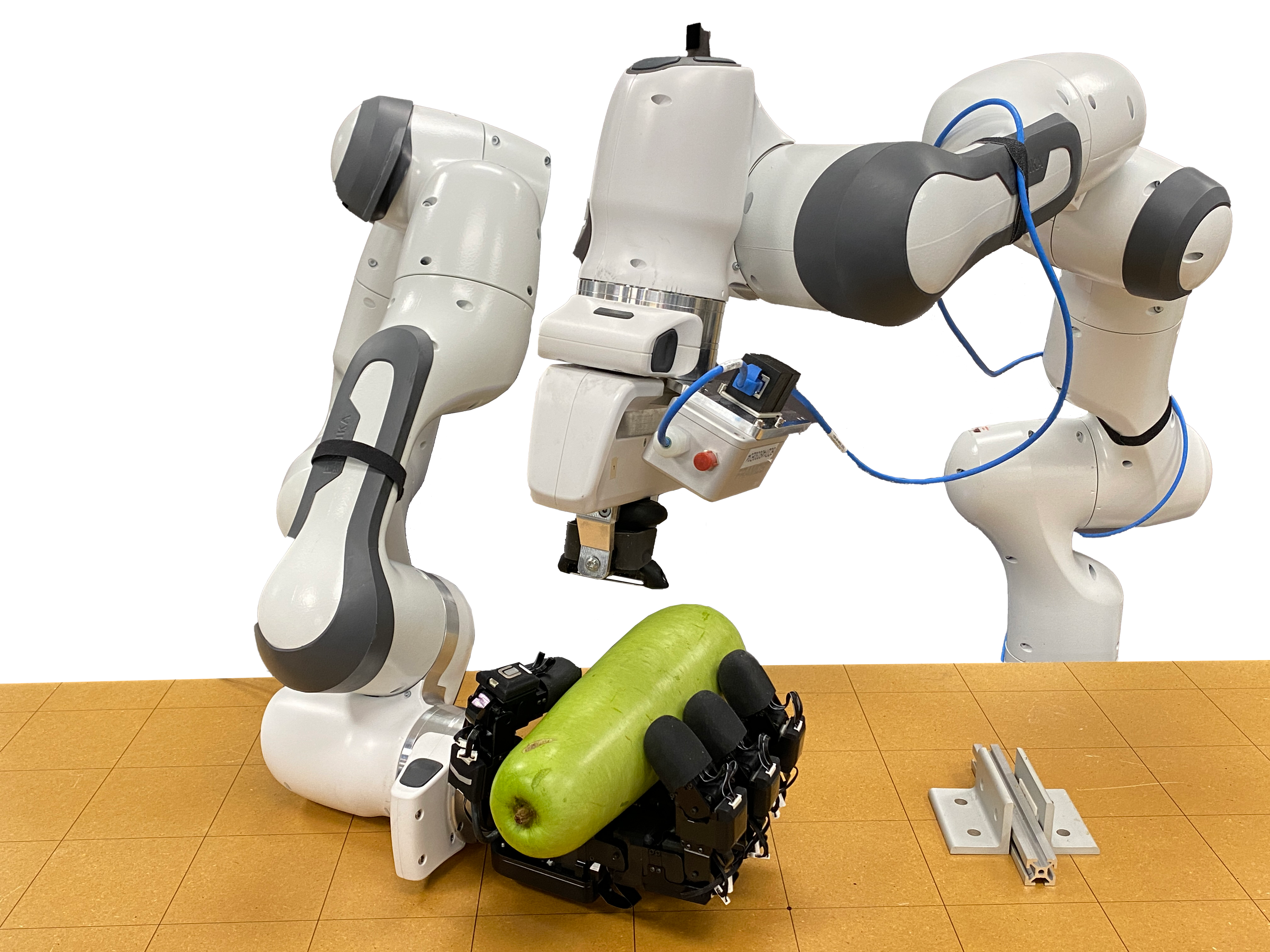}
  \end{center}
  \caption{Robot setup for reorientation and peeling.}
  \label{fig:peeling_setup}
\end{wrapfigure}

Having robots perform food preparation tasks has been of great interest in robotics. Imagine the scenario of making mashed potatoes, where a critical step is to peel potatoes. Humans peel potatoes by grasping the potato in one hand and using the second hand to actuate a peeler to remove the potato's skin. After a part of the potato is peeled, it is rotated while being held in the hand (i.e., \textit{in-hand manipulation}) and peeled again. The sequence of rotating and peeling continues until all of the potato's skin is removed. In this work, we present a robotic system that can re-orient different vegetables using an Allegro hand in a way that their skin can be peeled using another manipulator. Our setup is shown in \figref{fig:teaser} and \figref{fig:peeling_setup}.

In-hand rotation of vegetables is an instance of dexterous manipulation problem~\cite{rus1999hand}, a family of tasks that involves continuously controlling the force on an object while it is moving with respect to the fingertips~\cite{mason1989robot,dafle2014extrinsic}. The challenges in dexterous manipulation stem from the frequent making and breaking of contact, issues in contact modeling, high-dimensional control space, perception challenges due to severe occlusions, etc. A body of work made simplifying assumptions such as manipulating convex objects~\cite{bai2014dexterous,sundaralingam2019relaxed,rus1999hand,mordatch2012contact}, small finger motions\cite{pang2022global,morgan2022complex,abondance2020dexterous}, slow or quasi-static motion or manipulating a few specific objects~\cite{nagabandi2020deep,pang2022global,morgan2022complex} to leverage trajectory optimization or planning-based methods to achieve in-hand object re-orientation~\cite{rus1999hand,pang2022global,morgan2022complex,abondance2020dexterous,mordatch2012contact, bai2014dexterous, sundaralingam2019relaxed,nagabandi2020deep}.
Another line of work has used reinforcement learning for in-hand re-orientation\cite{chen2021system,chen2022visual,handa2022dextreme,yin2023rotating,khandate2022feasibility} and recent works have leveraged simulation to train policies capable of dynamically re-orienting a diverse set of new objects in real-time and in the real world~\cite{chen2021system,chen2022visual}. 

There are several challenges in adapting re-orientation controllers for a downstream task such as peeling vegetables. These challenges stem from the fact that controllers optimized for re-orientation~\cite{andrychowicz2020learning,handa2022dextreme,yin2023rotating,khandate2022feasibility,chen2022visual} are only optimized to continuously reorient the object and not to satisfy numerous constraints arising from task-specific requirements. For instance, peeling vegetables requires the hand to \textbf{first} \textit{stop} re-orienting the object and then for the peeler to peel the vegetable. Many prior works solve a version of the re-orientation problem where the object is continuously rotated ~\cite{qi2023hand,andrychowicz2020learning,handa2022dextreme} or otherwise perform quasistatic re-orientation~\cite{morgan2022complex}. Stopping and re-starting \textit{dynamic} re-orientation is difficult due to the challenge of dealing with the object's inertia. \textbf{Second}, the hand needs to \textit{hold} the object firmly enough to resist forces applied by the peeler. The closest work that attempts to hold the object at a target configuration~\cite{chen2022visual} is only able to loosely hold the object which is insufficient for resisting forces.  \textbf{Third}, the hand needs to reorient the vegetable \textit{along a specific axis in place}. Here, the specific axis refers to the rotational axis on the object that is parallel to the peeling direction. Similar to how humans reorient vegetables for peeling, it is desirable for the hand to reorient the object in place so that multiple consecutive cycles of reorientation and peeling can be performed. If the object substantially shifts its position during reorientation, the controller will struggle to reorient and hold the object at future time steps. \textbf{Fourth}, when the vegetable is held stationary the fingers should \textit{not obstruct} the top surface of the vegetable to ensure that the peeler can peel the vegetable.  
 
 While in-hand object reorientation has been widely studied~\cite{chen2021system,chen2022visual,andrychowicz2020learning,allshire2022transferring,handa2022dextreme,qi2023hand}, no prior works can satisfy the constraints mentioned above. Yet, these constraints become critical for downstream dexterous manipulation beyond object re-orientation. We use vegetable peeling as a case study to investigate the challenges and solutions for building a dexterous manipulation system that can operate under constraints. We develop a framework where we leverage reinforcement learning in simulation to train a policy that can perform object re-orientation under constraints. For the peeling task, we explored two approaches - a teleoperation-based method leveraging human guidance as well as an autonomous vision-based technique. Our contributions are as follows:
\begin{enumerate}
\item A framework for solving dexterous manipulation problems under the aforementioned constraints.
\item We propose a method that can make RL policy learn to stop its motion and hold objects firmly in hand -- a critical behavior for many downstream dexterous manipulation problems.
\item We present a step towards a robotic system capable of peeling diverse vegetables with different shapes, masses, and material properties while holding and manipulating the vegetables in hand.
\end{enumerate}

\section{Related Work}

\textbf{In-hand Object Reorientation}: Dexterous manipulation involves the use of high degrees-of-freedom (DoF) manipulators for object manipulation~\cite{okamura2000overview}. Its requirement for high-dimensional real-time control and its nature of frequent contact-making and breaking present grand challenges to roboticists. Recently, there has been a growth of interest in a particular instance of dexterous manipulation problems: in-hand object reorientation. This problem is of particular interest as it is a necessary step in many tool-use scenarios. For example, to use a screwdriver for tightening a screw, one has to reorient the screwdriver to align it with the screw. We can cluster the works in in-hand object reorientation from many aspects. For example, from the perspective of sensory information, \cite{bhatt2022surprisingly} studies open-loop cube reorientation without using any sensors, \cite{kumar2016learning, sundaralingam2019relaxed, andrychowicz2020learning,nagabandi2020deep,calli2018learning} use motion capture system or special tracking markers for object reorientation, \cite{qi2023hand} uses proprioceptive sensors such as joint encoders, \cite{ishihara2006dynamic,van2015learning,khandate2022feasibility,yin2023rotating} use tactile sensors and \cite{calli2017vision,andrychowicz2020learning,chen2022visual,allshire2022transferring} utilize vision sensors. In terms of the dynamics of the system, \cite{pang2022global,morgan2022complex,abondance2020dexterous} achieved object reorientation under the assumption of quasi-static motion where object moves slowly and its inertia effect can be ignored, while \cite{khandate2022feasibility,andrychowicz2020learning,chen2022visual,yin2023rotating,furukawa2006dynamic} focuses on dynamic object reorientation where object is manipulated in a fast and dynamic way. To make in-hand object manipulation useful for downstream tool use tasks, one important aspect of the skill is the ability of stably and firmly holding the object in end of the policy rollout. While many prior works on dynamic manipulation such as \cite{andrychowicz2020learning,nagabandi2020deep,yin2023rotating,khandate2022feasibility,qi2023hand} only consider endlessly rotating the object in hand and cannot stop the object stably when the object reaches the goal orientation, some works such as \cite{chen2022visual,furukawa2006dynamic} try to develop controllers that can reorient objects in hand and also hold the object in the goal orientation. Our work studies dynamic in-hand object manipulation with the capability of stopping objects stably in hand.

\textbf{Reinforcement Learning for Contact-rich Tasks}: Contact-rich tasks are particularly challenging due to the difficulty in modeling the system dynamics, especially when the tasks are performed in the wild, outside of a constrained and controlled setting. Examples of such tasks include quadruped robots hiking in mountains and robot hands reorienting various everyday objects. There have been many works using reinforcement learning to learn controllers for solving contact-rich tasks~\cite{akkaya2019solving, andrychowicz2020learning,handa2022dextreme,tan2018sim,da2021learning,li2021reinforcement,pitz2023dextrous}. In the real world, robots typically only have access to a limited amount of state information of the system due to the lack of sensors or the challenges in setting up the sensors. Using reinforcement learning to learn controllers from scratch with limited sensory information tends to be data-inefficient. One way to speed up policy learning is to provide asymmetric information to the policy and value function, where the value function observes much more privileged information~\cite{andrychowicz2020learning,handa2022dextreme,akkaya2019solving,pinto2017asymmetric}. Another method is to decouple policy learning into two stages: a reinforcement learning stage where agents (teacher) observe privileged fully-observable state information, and an imitation learning stage where the policy with limited sensory observation input (student) learns to imitate the policy with fully-observable state information. This approach has been successfully applied to various contact-rich problems such as locomotion~\cite{margolis2022rapid,margolis2021learning,li2021reinforcement,lee2020learning,kumar2021rma} and dexterous manipulation~\cite{chen2021system,chen2022visual,qi2023hand}. Our pipeline is built upon the idea of teacher-student policy learning and has made several key improvements, which we will detail below.

\section{Method}
\label{sec:method}
Peeling requires a reorientation controller that can stop its motion and firmly hold objects after reorientation. The first step in stopping is to decide when re-orientation should be stopped. One possibility is to have a perception system predict the desired rotation angle after which the next round of peeling would be performed. To accomplish the goal, the robot would need to track changes in object pose and compare it with the target rotation angle.  
However, accurately estimating object pose is challenging, especially when generalization to new objects is necessary~\cite{tremblay2018deep,andrychowicz2020learning,handa2022dextreme,pitz2023dextrous}.

One of our insights is that instead of training a predictor for desired rotation angle and object pose estimation, it can be \textit{easier} and \textit{sufficient} to train a \textit{binary vision classifier} that detects in real-time when the peeled part has been turned over. With such a classifier, the reorientation controller's job is simply to keep reorienting the object until it receives a stop signal. In this formulation, unlike prior works~\cite{chen2021system,chen2022visual}, the reorientation controller is not conditioned on target orientation but rather on a stop signal. Formally, the policy takes as input a binary variable $\stopsign\in\{0,1\}$ representing the stop signal. If $\stopsign=1$, the policy should stop immediately and ensure the fingers stably and firmly hold the object. Otherwise, the policy should continue reorienting the object. Note that in this work, we focus on learning the reorientation controller, leaving integration of a vision classifier to future work.

The next question is how to train such a policy. Using RL to train the policy from scratch can be challenging and requires extensive reward shaping because $\stopsign=1$ is a rare event in an episode, and when the $\stopsign$ is flipped to one from zero, the policy needs to quickly stop the motion posing a hard-exploration challenge.  

Prior works~\cite{chen2021system,chen2022visual} show success in training a goal-conditioned object reorientation controller. Can we leverage a goal-conditioned reorientation controller to train a controller that reacts to a stop signal? It turns out we can formulate this using the teacher-student learning framework~\cite{chen2021system,chen2022visual,chen2020learning,lee2020learning,margolis2021learning}. Specifically, we can use RL to train a goal-conditioned controller that reorients an object by random goal angles along its rotational axis. This acts as the teacher. Next, we can use imitation learning (specifically DAGGER~\cite{ross2011reduction}) to train a controller conditioned on the stop signal to imitate the teacher. The stop signal can be generated during training by checking if the orientation distance to the goal is below a threshold. Using imitation learning bypasses the hard exploration challenge.

\subsection{Teacher Policy Learning: Reorient and Stop}
We train the teacher policy to re-orient the object along a pre-defined axis and stop (see \figref{fig:rotation_illustration}). The teacher is formulated as a goal-conditioned policy $\bm{a}^\expsymbol_t=\expert(\obexpert_t, \bm{a}_{t-1}, g)$, where $\expsymbol$ represents variables for the teacher policy, $\bm{o}_t$ is the observation, $\bm{a}_t$ is the action command, $g$ is the goal representing the amount by which the object needs to be re-oriented. $g$ is randomly and uniformly sampled from $[1.57, 4.0]$rad during training. 

While the teacher policy's formulation is similar to that in prior works~\cite{chen2021system,chen2022visual}, we propose (i) a much simpler reward function, (ii) new success criteria that effectively encourages the policy to stop the object and firmly hold it, and (iii) an interpolation scheme that enables smoother policy actions in the real world.

\subsubsection{Reward Function}
A common approach to designing the reward function is to create multiple terms that make it easier for the manipulator to discover the desired behavior (i.e., reward shaping). For instance, to facilitate exploration, we can devise a reward term that reduces the distance between the fingertips and the center of mass (CoM) of the object. To discourage excessive translational motion of the object during rotation, we can create a reward term that penalizes the displacement of the CoM. To discourage the object from rotating with undesired motion along other axes, we can add another reward term that reduces the distance between the tip of the thumb and the centerline of the palm. This ensures that the thumb applies force close to the object's CoM, rather than to one side of the object. Additionally, we need to design a reward term that discourages the fingers from covering the top surface of the object, which affects peeling. Hence, designing multiple reward terms is necessary to regulate the behavior under specific constraints. Balancing these terms requires extensive hyper-parameter tuning.

For the task of in-hand re-orientation, we found that the reward function can be substantially simplified by using a task demonstration. However, unlike prior works that rely on trajectory-level demonstrations~\cite{ho2016generative,peng2021amp}, our method only requires a \textit{one-step demonstration} (a keyframe), which is much easier to collect. Specifically, we manually move the real Allegro hand to a good pose where the constraints mentioned above are satisfied (e.g., the fingers do not cover the food item), and the fingers touch the object and are ready to reorient it. We record the joint positions as $\bm{q}^{demo}$. During training in simulation, we encourage the joint positions at any time step to be close to $\bm{q}^{demo}$.

Overall, our reward function is as follows:
\begin{align}
    r_{t} = c_1 \mathds{1}(\text{Task successful}) + c_2\frac{1}{|\Delta \theta_t|+\epsilon_\theta}
    + c_3\left\Vert \bm{q}_t-\bm{q}^{demo}\right\Vert_2^2 
\end{align}

where $c_1=800, c_2=1.5, c_3=-0.6$ are coefficients. $\mathds{1}(\text{Task successful})$ is $1$ when the task is successfully completed, and $0$ otherwise. $\Delta \theta_t$ is the distance between the object's current and goal orientation. The first two terms are task rewards for object reorientation. The last term is to regulate hand behavior.

\subsubsection{Success Criteria}
In a goal-conditioned object reorientation, a common way to claim the task successful is by checking if the distance between the object's current and the goal orientation is smaller than a threshold value (orientation criterion $C_{ori}=\Delta\theta<\Bar{\theta}$) \cite{andrychowicz2020learning,handa2022dextreme}. Another criterion is that all the fingertips should make contact with the object (contact criterion $C_{contact}$), a pre-requisite for firmly holding the object after reorientation. However, only checking these two criteria is insufficient to ensure the policy learns to stop the motion and hold the object firmly around the goal orientation, as discussed in \cite{chen2022visual}. The policy can oscillate around the goal state due to observation and control delay and noise.

To further encourage the policy to stop robot motion when the goal is reached and firmly hold the object, we propose adding time constraints to the success criteria: both $C_{ori}$ and $C_{contact}$ should be continuously satisfied for $\Bar{T}^{succ}$ time steps. Adding this criterion makes the MDP partially observable since the policy's observation lacks the knowledge of time. Therefore, to facilitate policy learning, we augment the observation space with a scalar indicator variable $I^{succ}= t^{succ}/\Bar{T}^{succ}\in [0, 1]$, where $t^{succ}$ is the number of consecutive steps satisfying $C_{ori}$ and $C_{contact}$. The observation space becomes $\obexpert:=\obexpert\oplus I^{succ}$. In this work, $\bar{\theta}=0.2$rad, $\Bar{T}^{succ}=8$.

\subsubsection{Reset Constraints}
\begin{figure}[t!]
\centering

\begin{subfigure}{0.2\linewidth}
\centering
\includegraphics[width=\linewidth]{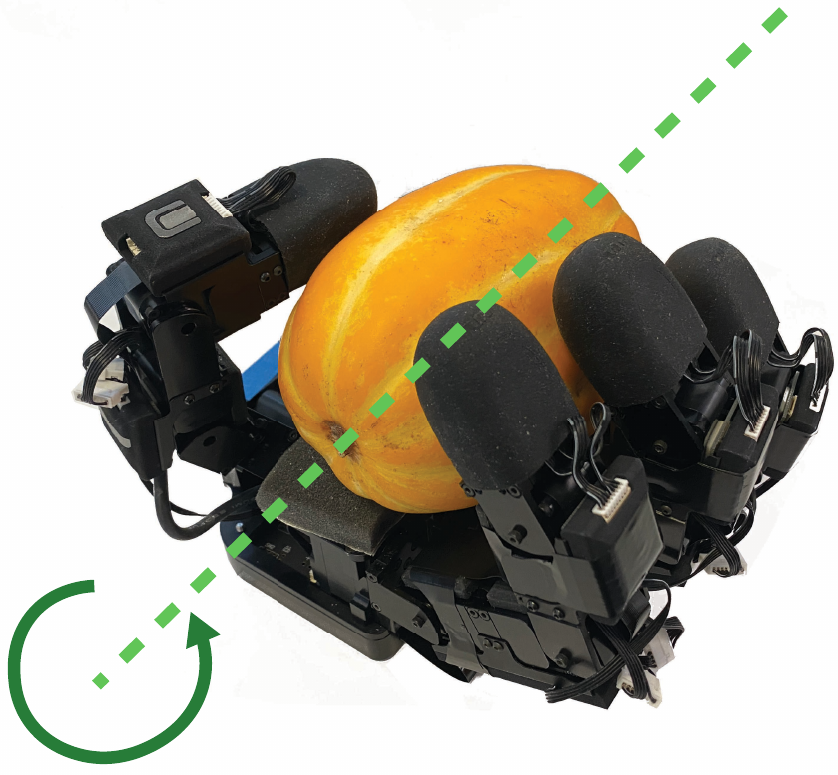}
\caption{}
\label{fig:rotation_illustration}
\end{subfigure}%
\hfill
\begin{subfigure}{0.2\linewidth}
\centering
\includegraphics[width=\linewidth]{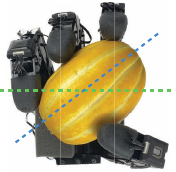}
\caption{}
\label{fig:big_deviation}
\end{subfigure}%
\hfill
\begin{subfigure}{0.3\linewidth}
\centering
\includegraphics[width=\linewidth]{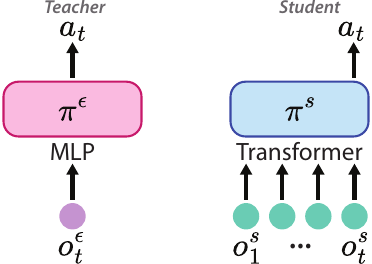}
\caption{}
\label{fig:net_arch}
\end{subfigure}
\caption{\textbf{(a)} shows an example of the rotational axis of a melon. \textbf{(b)} shows an example where the object's orientation (the blue line) has a large deviation from the desired rotational axis (the green line). We reset the episode when this occurs. \textbf{(c)} shows the policy Architecture for the teacher and the student. In this figure, we use $\bm{o}_t$ to represent all the policy input at each time step.}
\label{fig:setup}
\vspace{-0.5cm}
\end{figure}

As mentioned earlier, a reorientation policy for peeling needs to meet several constraints, such as in-place and fixed-axis reorientation (\figref{fig:big_deviation}). While one could design individual reward terms to satisfy these constraints, tuning these reward terms to achieve the desired result can be difficult. Instead, it is much simpler to formulate the constraints as reset conditions. In other words, if the constraints are violated, the episode is reset immediately. This incentivizes the policy to explore only in space where the constraints are satisfied. Similar techniques were also used in some prior works~\cite{chen2021system,chen2022visual,yin2023rotating}.

\subsubsection{Interpolation and Reference for Action Commands}
Our neural network controller operates at a relatively low control frequency of $12$Hz. To track the joint position command, a low-level PD controller runs at $300$Hz. To ensure smoother joint motion, we interpolate the low-frequency joint position commands. While more complex interpolation schemes such as spline interpolation are possible, we found that simple linear interpolation is sufficient to generate smooth higher-frequency ($60$Hz) joint position commands. To do this, we linearly interpolate between the current reference joint positions ($\bm{q}_t^{ref}$) and the desired joint positions ($\bm{q}_{t+1}^{cmd}$) for the next policy control time step. We then send the interpolated joint position commands to the PD controllers. Mathematically, $\bm{q}_{t+1}^{cmd,n}=\bm{q}_t^{ref}+\frac{n}{N}\bm{a}_t$, where $n\in[1, N]$ ($N=5$) and $\bm{q}_{t+1}^{cmd,n}$ represents the $n^{th}$ interpolated joint position command for the next policy control time step.

\begin{figure}[htbp]
\centering
\begin{subfigure}{0.3\linewidth}
\centering
\includegraphics[width=\linewidth]{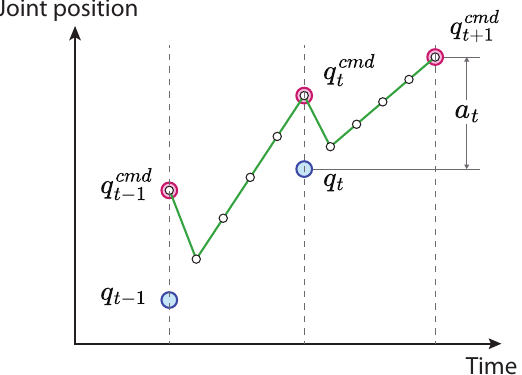}
\caption{}
\label{fig:q_interp}
\end{subfigure}
\qquad
\begin{subfigure}{0.3\linewidth}
\centering
\includegraphics[width=\linewidth]{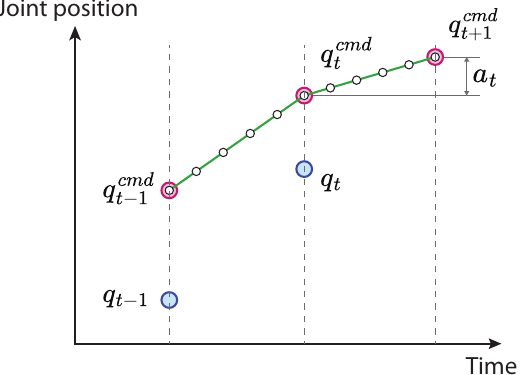}
\caption{}
\label{fig:q_tgt_interp}
\end{subfigure}
\caption{Examples of joint position commands after interpolation sent to a low-level PD controller. \includegraphics[width=\charwidth]{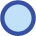} represents the actual joint position of the motor. \includegraphics[width=\charwidth]{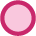} is the computed desired joint position. \includegraphics[width=\charwidth]{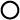} on the green line shows the interpolated joint position commands that are sent to the low-level PD controller. \textbf{(a)} shows the case of $\bm{q}_{t+1}^{cmd}=\bm{q}_t+\bm{a}_t$, while \textbf{(b)} shows the case of $\bm{q}_{t+1}^{cmd}=\bm{q}_t^{cmd}+\bm{a}_t$. We can see that \textbf{(b)} generates much smoother joint commands.}
\label{fig:interp}
\end{figure}

When the action space is chosen as the change in joint position, the target joint position for the PD controller is calculated as follows: $\bm{q}_{t+1}^{cmd}=\bm{q}_t+\bm{a}_t$~\cite{chen2022visual,chen2021system,andrychowicz2020learning}. Here, $\bm{q}_t$ is the current joint position, and $a_t = \Delta\bm{q}_t$ is the desired change in joint positions, as described earlier. In this case, the reference is chosen to be the current joint positions, i.e., $\bm{q}_t^{ref}=\bm{q}_t$. However, we found that this scheme results in significant jerky motion when combined with action interpolation. To illustrate this, consider a simplified example of one joint, as shown in \figref{fig:q_interp}. Since we are using a PD controller only to control the joint position, there is usually an error in tracking the joint position command, as shown by the difference between $q_t^{cmd}$ and $q_t$. If we set $q^{ref}_t=q_t$, when we interpolate between $q^{ref}_t$ and $q^{cmd}_{t+1}$, it tends to cause a sudden change in the PD controller's set point, as shown in ~\figref{fig:q_interp}. A sudden change in the set point can cause a sudden change in the joint torque command and hence cause jerky motion. To resolve this issue, we use the previous joint position command as the reference, as shown in \figref{fig:q_tgt_interp}. In other words, $\bm{q}^{ref}_t=\bm{q}_t^{cmd}$, and $\bm{q}_{t+1}^{cmd}=\bm{q}_t^{cmd}+\bm{a}_t$.

\subsection{Student Policy Learning: Imitate and Stop}
After learning a goal-conditional teacher policy $\bm{a}^\expsymbol_t=\expert(\obexpert_t, \bm{a}_{t-1}, g)$, the next question is how to train a real-world deployable student policy that can rotate the object in hand and hold it stably after reorientation. We propose conditioning the student policy on a stop signal $\stopsign\in\{0, 1\}$: $\bm{a}^\stusymbol_t=\student(\obstudent_t, \bm{a}_{t-1},\stopsign)$. In other words, the student policy should continue reorienting the object when $\stopsign=0$, but stably hold the object when $\stopsign=1$. This design choice provides flexibility in how we control the policy to stop the reorientation. For example, the policy could rotate the object for a pre-specified amount of time (i.e., set $\stopsign=1$ after $t$ seconds). Alternatively, an external perception module could detect when the peeled part has fully turned over, triggering $\stopsign=1$ and the policy to stop the motion and hold the object immediately.

How can we use the learned goal-conditioned teacher policy to train a student policy that is conditioned on the stop signal? We can set the value for $\stopsign$ automatically during policy rollout based on the orientation distance $\Delta\theta_t$.
\begin{align*}
    I_t^{stop}= \begin{cases}
   0 & \text{if $\Delta\theta_t>\Bar{\theta}$} \\
   1 & \text{otherwise}
\end{cases}
\end{align*}

Details about the observation space and the policy architecture are in \secref{subsec:student} in the appendix.

\subsection{Peeling}
In this section, we demonstrate that our reorientation controller can be used for downstream peeling tasks. We use the dexterous robot hand to do the reorientation and then control another Franka Panda robot arm to do the peeling as shown in ~\figref{fig:peeling_setup}. To control the robot arm, we experimented with both using a teleoperation system and an automatic vision-based peeling system.

\subsubsection{Teleoperation-based peeling}
We used a leader-follower teleoperation system in which a human operator controls a leader system, and the Franka arm follows the motion of the leader in real-time. A 200 Hz operational space impedance controller~\cite{khatib1987osc} runs on the Panda arm, controlling for pose via torque, and an operator interacts with a Haption Virtuose\legalTM\ 6D HF TAO%
\footnote{\url{https://www.haption.com/en/products-en/virtuose-6d-tao-en.html\#fa-download-downloads}} device. Bilateral position-position haptic coupling is done between the two devices. The controllers and haptic coupling are implemented using Drake~\cite{drake}. 

\begin{figure}[t!]
    \centering
    \includegraphics[width=\linewidth]{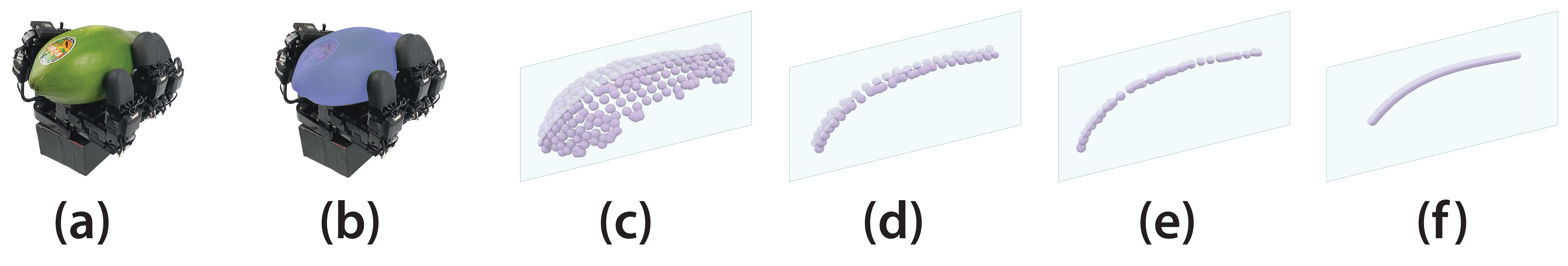}
    \caption{\textbf{(a)}: the Allegro hand holds a papaya to be peeled. \textbf{(b)}: we utilize Grounded SAM to segment the papaya. \textbf{(c)}: the 3D point cloud representing the segmented papaya's exposed surface. \textbf{(d)}: we take a slice of this point cloud at the center region along the papaya's longest axis. \textbf{(e)}: the points within this center slice are projected onto the central plane aligned with the axis. \textbf{(f)}: we fit a spline curve to the projected points to obtain the desired trajectory for the peeler tip to follow.}
    \label{fig:peeling_vision}
    \vspace{-0.5cm}
\end{figure}
\subsubsection{Vision-based peeling}
While teleoperation provides effective peeling commands for the Franka arm and demonstrates that our reorientation controller can firmly grasp objects after reorientation, automating the peeling process would be ideal. One approach to achieve this is by computing the peeler's motion trajectory based on RGB and depth vision data. The trajectory can be determined through the following steps (see \figref{fig:peeling_vision}): (1) We utilize Grounded SAM~\cite{ren2024grounded} to segment the target vegetable given an image and vegetable name input. (2) Using the segmentation mask and depth data, we reconstruct the 3D point cloud representing the vegetable's top surface. (3) We identify the vegetable's longest axis (the peeling direction) by applying principal component analysis. (4) We slice the point cloud into a 2cm thick segment along the central plane that crosses the center point and aligns with the longest axis. We then project all the points within the slice onto the plane. (5) We fit a spline curve to the projected points to obtain a smooth trajectory for the peeler tip. Finally, cartesian-space position control moves the peeler along this trajectory while keeping the peeler orientation fixed.

\section{Results}

To quantitatively evaluate the real-world policy transfer performance, we tested the controller on four vegetables (\figref{fig:real_objects}): a pumpkin (mass: $827$g), a melon($623$g), a radish($727$g), a papaya($848$g).

\subsection{Traveling distance for a fixed amount of commanded motion time}
The first question we want to answer is whether the learned policy can successfully reorient vegetables in the real world. In peeling, the width of the peeled part depends on the peeler's width. Thus, it is more informative to measure how much the reorientation controller rotates an object by the traveling distance of a surface point, rather than the absolute rotation angle. Specifically, we mark a reference point $P^{ref}$ on the object surface near the mid-point of its rotational axis. At the start, we ensure $P^{ref}$ is centered and facing upward when held. After reorientation, we record the new point $P^{new}$ that is now centered and facing upward. We then measure the contour length from $P^{new}$ to $P^{ref}$ along the surface (\figref{fig:travel_dist_illustration}).

\begin{figure}[!tbp]
\centering
\begin{subfigure}{0.35\linewidth}
\centering
\includegraphics[width=\linewidth]{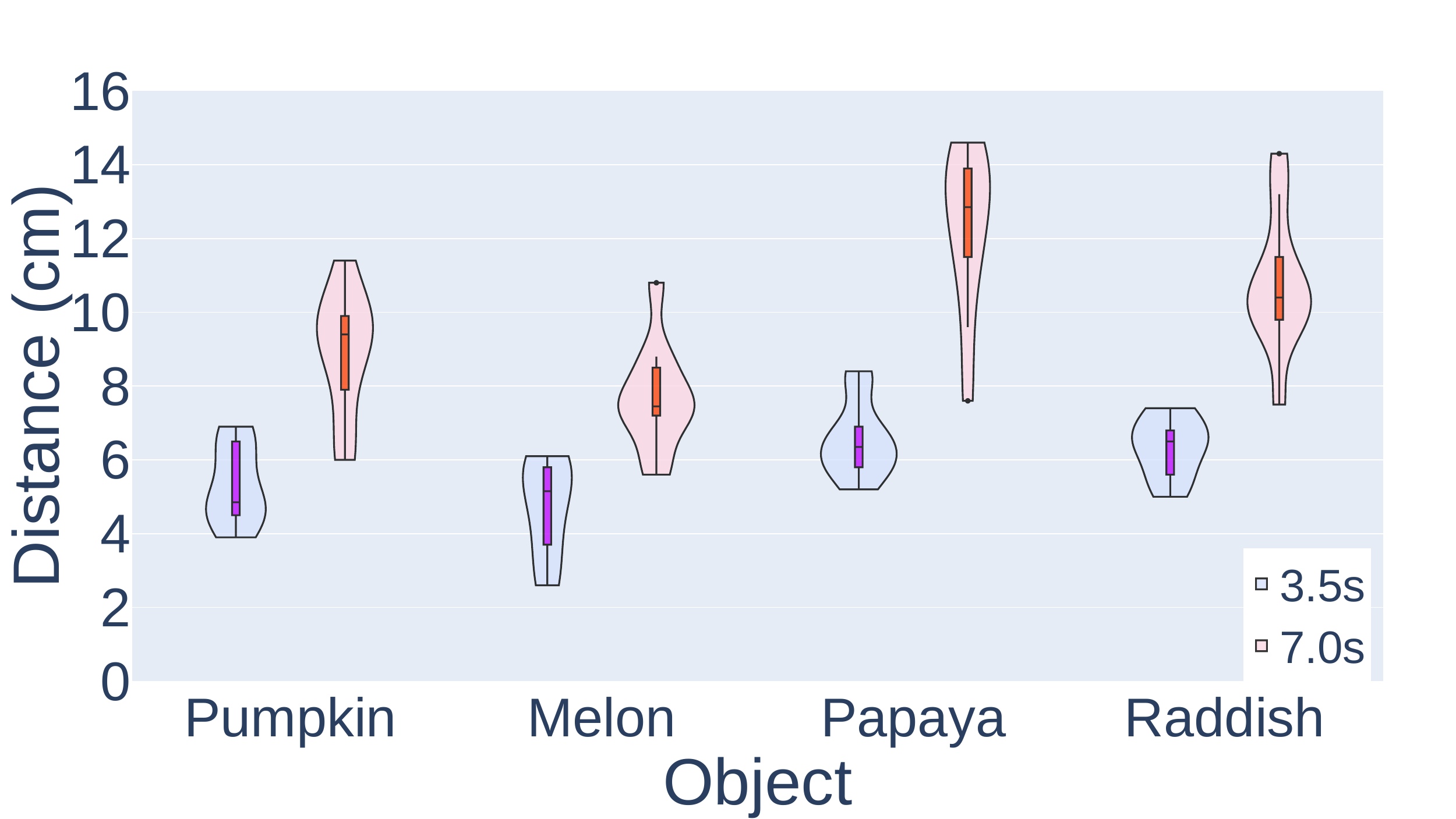}
\caption{}
\label{fig:travel_dist}
\end{subfigure}
\begin{subfigure}{0.35\linewidth}
\centering
\includegraphics[width=\linewidth]{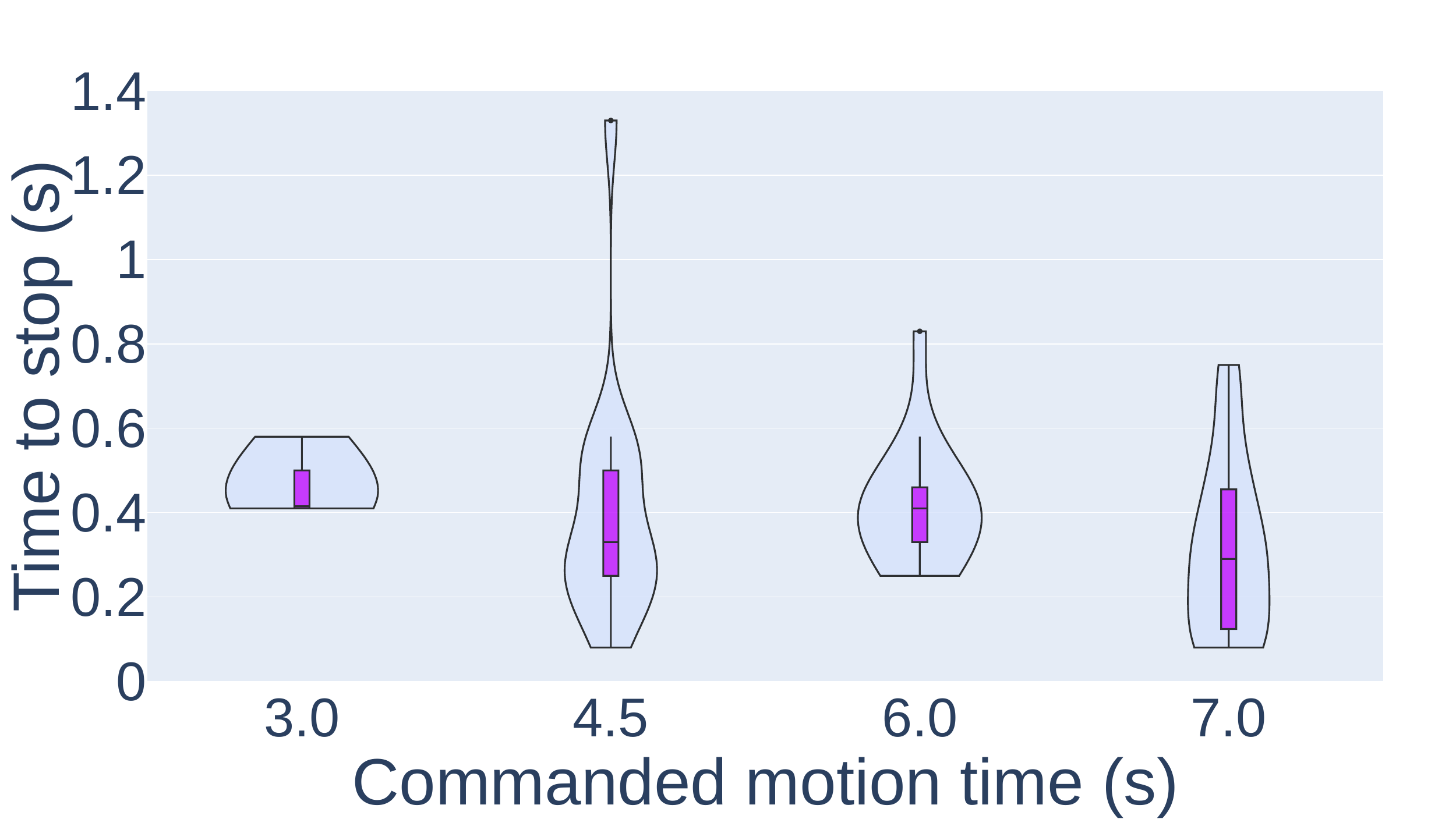}
\caption{}
\label{fig:time-response}
\end{subfigure}
\caption{\textbf{(a)}: Violin plots showing the distribution of the traveling distance of a point on the object surface after the controller is commanded to rotate the object for 3.5 s and 7 s, respectively. \textbf{(b)}: Violin plot showing the distribution of time taken by the controller to transition from rotating the object in hand to firmly holding the object after receiving the stop signal. The $x$-axis represents the timing of the stop signal sent to the controller after it starts.}
\label{fig:violin}
\vspace{-0.5cm}
\end{figure}

To demonstrate the capability of our controller to reorient real objects, we conducted two rounds of testing. Our controller is trained to stop motion when it receives a stop signal. In the first round, we sent the stop signal 3.5 seconds after the controller started rotating. In the second round, we sent the stop signal 7 seconds after start. We repeated each test 10 times. As shown in \figref{fig:travel_dist}, the controller successfully reoriented all four food items by a sufficient amount for peeling. When commanded to reorient for 3.5s, 90\% of tests reoriented the objects by at least 4cm. With 7s, 90\% of tests reoriented objects by at least 7.3cm. Given more time, the controller reoriented objects by a larger amount.

\subsection{How well does the controller track the commanded motion time?}
\vspace{-0.2cm}
As discussed in Section~\ref{sec:method}, if our controller can quickly respond to a stop signal at any time step, it can be combined with a perception system that tracks peeling progress. Hence, we measured how long it takes to stop the hand and object motion after receiving the stop signal. As shown in \figref{fig:time-response}, the motion stops after 0.4s on average after the controller receives the stop signal.

\subsection{Firm grasp after reorientation}
\vspace{-0.2cm}
To enable downstream peeling, the reorientation controller must learn to firmly grasp the object after stopping finger motion. We tested this by checking if the Allegro hand and object could be lifted in the air for 3s by only lifting the object with a single human hand. \tblref{tbl:lift_succ} in the appendix shows that across objects and commanded times, the controller firmly grasped objects in 90\% of tests. Moreover, our controller possesses the capability of performing consecutive reorientations. It can repetitively execute the sequence of peeling and reorientation multiple times in succession.

\subsection{Real-world Peeling}
\vspace{-0.2cm}
We evaluated whether the reorientation controller could reorient food items to facilitate peeling (\figref{fig:teaser}). We tested using an Allegro hand and a Leap hand~\cite{shaw2023leap}. Testing showed that peeling applied substantial pulling forces on objects. However, in most cases, both hands maintained a firm enough grasp to enable successful peeling. Failures often occur when holding small objects, as some fingertips may fail to establish secure contact with the surface.

\section{Discussions}
The reorientation controller presented in this study is a blind controller that relies solely on proprioceptive sensory information. While it has demonstrated the ability to successfully reorient heavy objects and securely hold them in place, its performance could potentially be enhanced by incorporating visual and tactile feedback. The current system has a few failure modes. Firstly, the object might slip out of the hand since the controller does not utilize any vision information. Secondly, the controller might fail if the vegetables are small, as the fingers cannot effectively make contact with the object. When using a vision-based peeling approach to peel the vegetables, the segmentation network (Grounded SAM) might fail to correctly identify and segment the target vegetable in the image. Sometimes, the segmentation mask would incorrectly include the robot hand. Some fine-tuning of the pre-trained Grounded SAM model would be necessary to mitigate such issues. Future work could involve learning a peeling policy via behavior cloning on data collected via teleoperation to achieve better autonomy of the system. Additionally, incorporating visual and tactile feedback into the reorientation controller could potentially enhance its performance


\clearpage
\acknowledgments{We thank the anonymous reviewers for their helpful comments in revising the paper. We also extend our appreciation to the members of Toyota Research Institute for their valuable feedback on the formulation of our research idea and their engaging discussions about related research problems.}


\bibliography{references}  
\newpage
\begin{appendices}
\setcounter{figure}{0}
\setcounter{table}{0}
\setcounter{footnote}{0}
\renewcommand{\thefigure}{\Alph{section}.\arabic{figure}}
\renewcommand{\thetable}{\Alph{section}.\arabic{table}}

\section{Training}

\subsection{Training setup}

\textbf{Robot}: We use an Allegro Hand that is controlled via a PD controller at $300$Hz. Our control policy sets joint position commands and runs at a lower frequency at $12$Hz. 

\textbf{Simulation}: We trained the policies in Isaac Gym simulation~\cite{makoviychuk2021isaac}. To set dynamics-related robot parameters in the simulation, we followed a prior approach~\cite{chen2022visual}, which uses a gradient-free search method to find the dynamics parameters for each joint (joint friction, damping, maximum joint velocity, and maximum effort) in simulation that generates the motor response that is closest to the real motors.

\textbf{Object Dataset}: We collected $23$ object meshes (potatoes, squash, cucumber, etc.) from Objaverse~\cite{deitke2023objaverse}. $10$ variants for each mesh were created by varying the size. The mass of the object was randomly sampled in the range of $[80, 960]$g. Note that we aim to reorient much heavier objects than prior works~\cite{andrychowicz2020learning,chen2022visual,chen2021system,handa2022dextreme}. 

\begin{figure}[!htb]
    \centering
    \includegraphics[width=\linewidth]{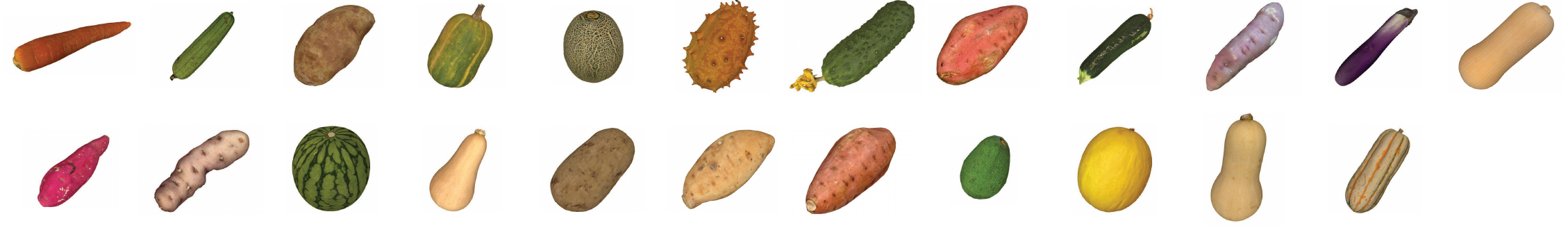}
    \caption{Object dataset used in this work. We collected meshes of carrot, sweet potato, potato, squash, pumpkin, etc.}
    \label{fig:objects}
\end{figure}

\subsection{Teacher Policy Learning}

\subsubsection{Observation and Action Space}
$\obexpert_t$ includes joint positions and velocities, the fingertip poses and velocities, object pose and velocity, the distance between the current object orientation and the goal orientation, and whether any of the fingertips touch the object. $\bm{a}_t$ is the delta joint position command. The neural network policy runs at $12$Hz.

\subsubsection{Domain randomization and Perturbation during training}
During training, we apply domain randomization on the joint stiffness and damping, friction, and restitution. Additionally, we randomly apply a perturbation force on the object's CoM. We randomly sample the direction of the perturbation force and set its magnitude to 10$m_o$, where $m_o$ is the object mass.

\subsection{Student Policy Learning}
\label{subsec:student}

\subsubsection{Observation Space} In this work, we only use proprioceptive sensory information (joint positions $\bm{q}_t$ and velocities $\dot{\bm{q}}_t$) as the observation input ($\obstudent_t$). Our findings indicate that relying solely on proprioceptive sensory information results in strong performance. Future research could investigate incorporating visual data to further enhance the system's capabilities, such as preventing objects from slipping out of the grasp.

\subsubsection{Policy Architecture} As the student policy only has access to a limited amount of sensory information (a POMDP setting), it is important to incorporate history information, as has been done in previous works~\cite{andrychowicz2020learning,handa2022dextreme,chen2022visual}. While \cite{andrychowicz2020learning,handa2022dextreme,chen2022visual} utilized RNNs to process history information, Transformers~\cite{vaswani2017attention} have gained significant attention due to their improved performance and faster training in domains such as natural language processing. Therefore, in this work, we employ a Transformer-based policy architecture. $\bm{a}^\stusymbol_t=\student(\obstudent_1, \bm{a}_{0},I_1^{stop}, ..., \obstudent_t, \bm{a}_{t-1},\stopsign)$. The policy is a decoder-only attention network (\figref{fig:net_arch}) with three self-attention layers. The hidden size is $256$, the intermediate size is $512$, and the number of attention heads is $8$. The policy is trained using DAGGER~\cite{ross2011reduction}.


\section{Testing}

\subsection{Testing setup}
\begin{figure}[t!]
\centering
\begin{subfigure}{0.4\linewidth}
\centering
\includegraphics[width=\linewidth]{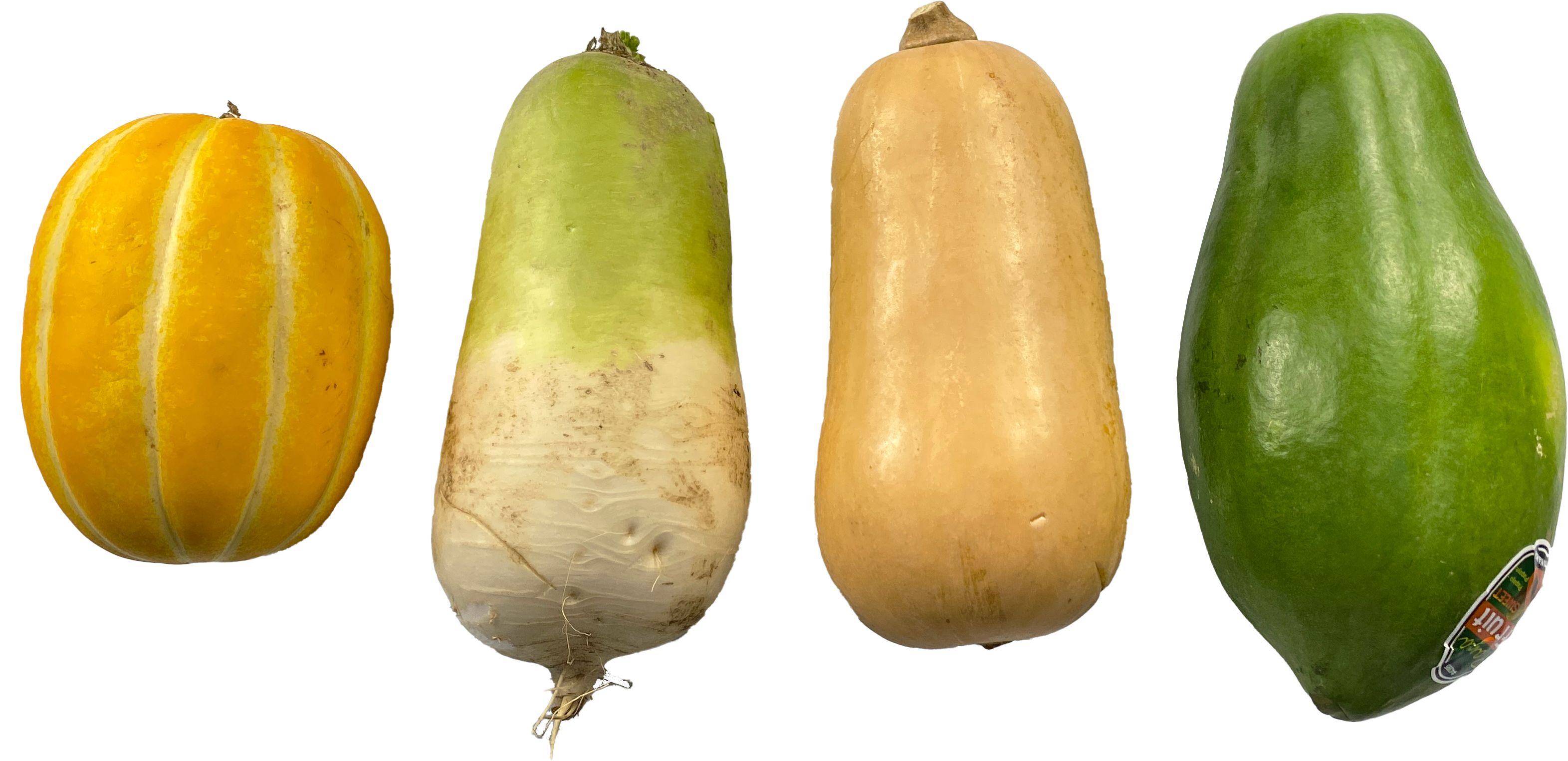}
\caption{}
\label{fig:real_objects}
\end{subfigure}
\qquad
\begin{subfigure}{0.3\linewidth}
\centering
\includegraphics[width=\linewidth]{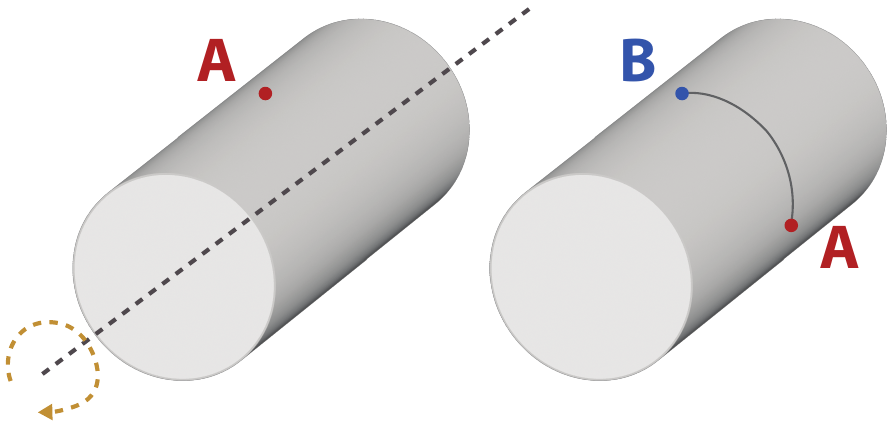}
\caption{}
\label{fig:travel_dist_illustration}
\end{subfigure}
\caption{\textbf{(a)} shows the objects for evaluation: melon, radish, pumpkin, papaya. \textbf{(b)} shows the traveling distance. Before reorientation begins, we ensure a reference point (point A) is facing upward. After reorientation, we identify the point (point B) now facing upward. We then measure the distance from point A to point B along the contour.}
\label{fig:obj_travel}
\end{figure}

\figref{fig:real_objects} show the objects used for evaluation. \figref{fig:travel_dist_illustration} illustrates how we measure the traveling distance of the rotation motion.

\subsection{Firm grasp after reorientation}

\tblref{tbl:lift_succ} shows the success rate of the lifting action after the reorientation. It shows that our reorientation controller can control the fingers to firmly hold the object after the reorientation.

\begin{table}[!tb]
\centering
\caption{Successful lifting rate (10 tests each)}
\label{tbl:lift_succ}
\begin{tabular}{ccccc}
\hline
Commanded motion time  & Pumpkin & Melon & Papaya & Radish \\ \hline
3.5s & 80\%    & 90\%  & 80\%   & 90\%    \\
7s   & 100\%   & 90\%  & 100\%  & 90\%    \\ \hline
     &         &       &        &        
\end{tabular}
\end{table}

\subsection{Ablation study}
\begin{figure}[t!]
\centering
\begin{subfigure}{0.45\linewidth}
\centering
\includegraphics[width=\linewidth]{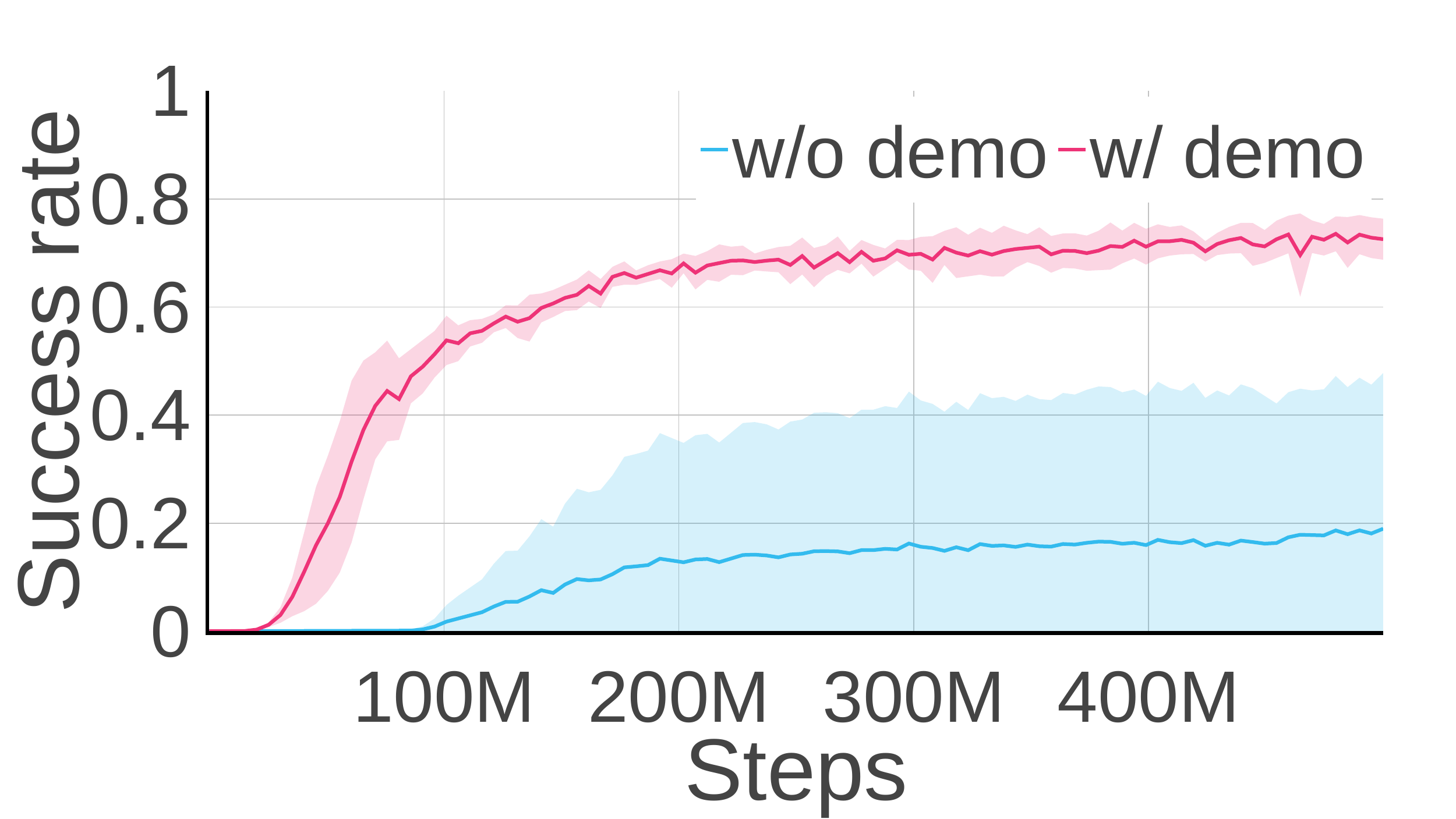}
\caption{}
\label{fig:ref_dof_step}
\end{subfigure}
\hfill
\begin{subfigure}{0.45\linewidth}
\centering
\includegraphics[width=\linewidth]{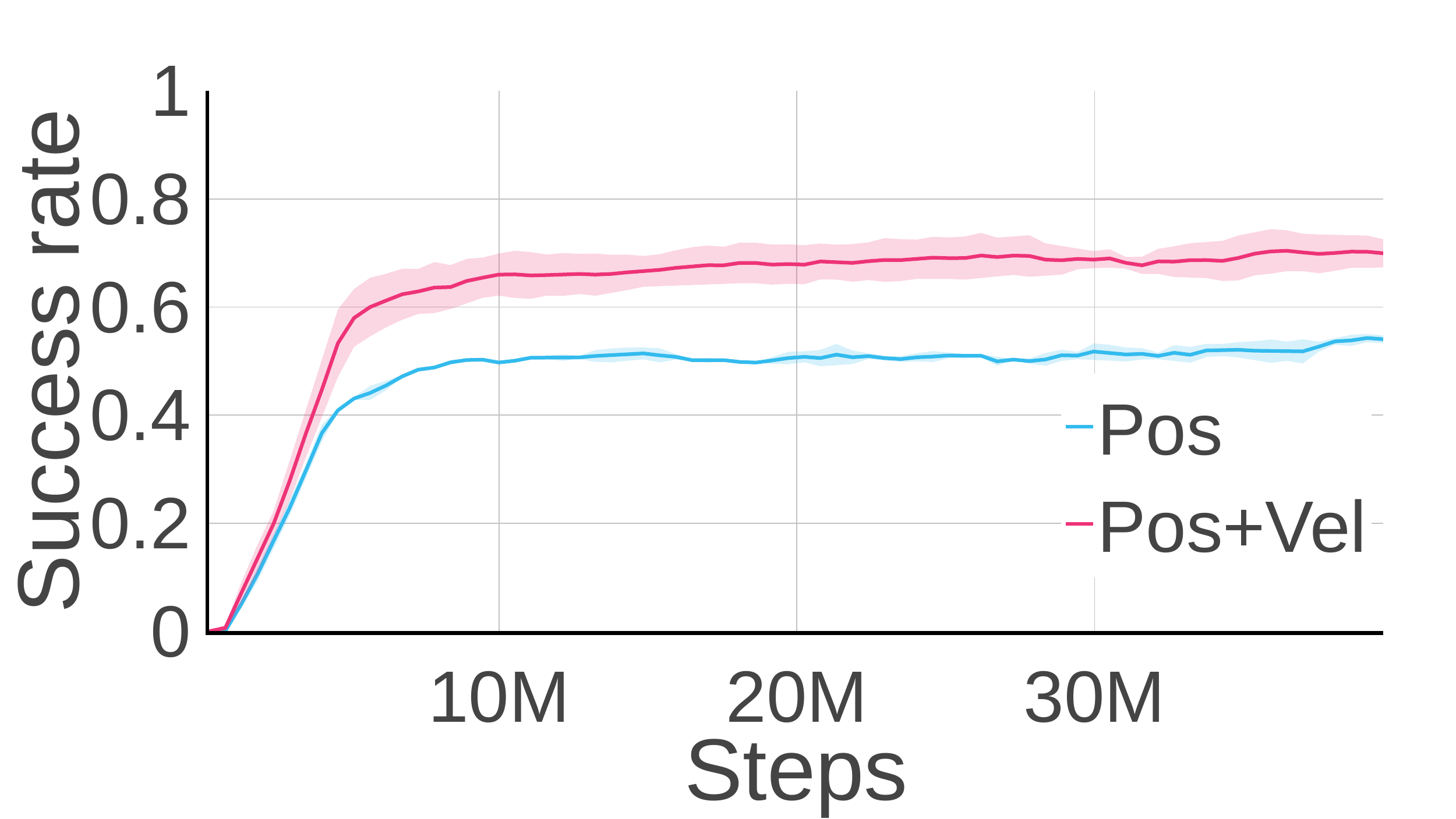}
\caption{}
\label{fig:ob_space_step}
\end{subfigure}
\caption{\textbf{(a)} shows learning curves of the teacher policies with or without $c_3\left\Vert \bm{q}_t-\bm{q}^{demo}\right\Vert_2^2$ in the reward function. \textbf{(b)} shows the differences between student policies trained with different sensory information (joint positions and velocities vs. joint positions only).}
\label{fig:demo}
\end{figure}

\paragraph{Demo term in Reward function}
We proposed using a keyframe demonstration to ease reward shaping. To evaluate its effectiveness, we compared learning curves of the teacher policies trained with and without the $c_3\left\Vert \bm{q}_t-\bm{q}^{demo}\right\Vert_2^2$ reward term. As shown in \figref{fig:ref_dof_step}, adding the keyframe substantially improved learning. Additionally, it demonstrates that mimicking the keyframe pose via a single reward term effectively reduces the reward-shaping burden.

\paragraph{Necessity of having joint velocity information in $\student$}

The student policy's sensory input included joint positions and velocities. We investigated whether including joint velocity information in the input is beneficial. \figref{fig:ob_space_step} shows that adding joint velocities to the input improved performance.

\begin{figure}[!htbp]
\centering
\begin{subfigure}{0.45\linewidth}
\centering
\includegraphics[width=\linewidth]{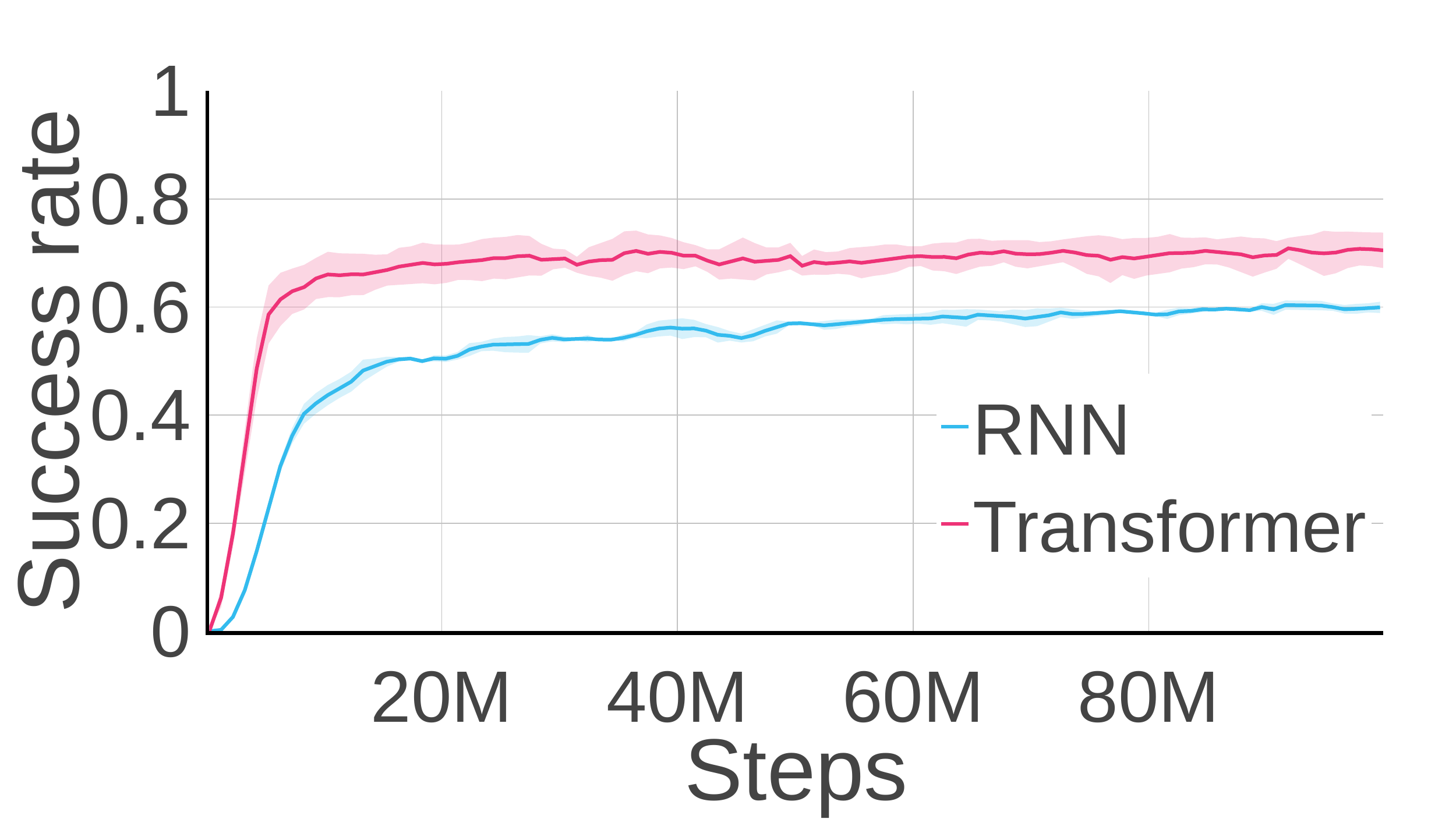}
\caption{}
\label{fig:transformer_step}
\end{subfigure}
\hfill
\begin{subfigure}{0.45\linewidth}
\centering
\includegraphics[width=\linewidth]{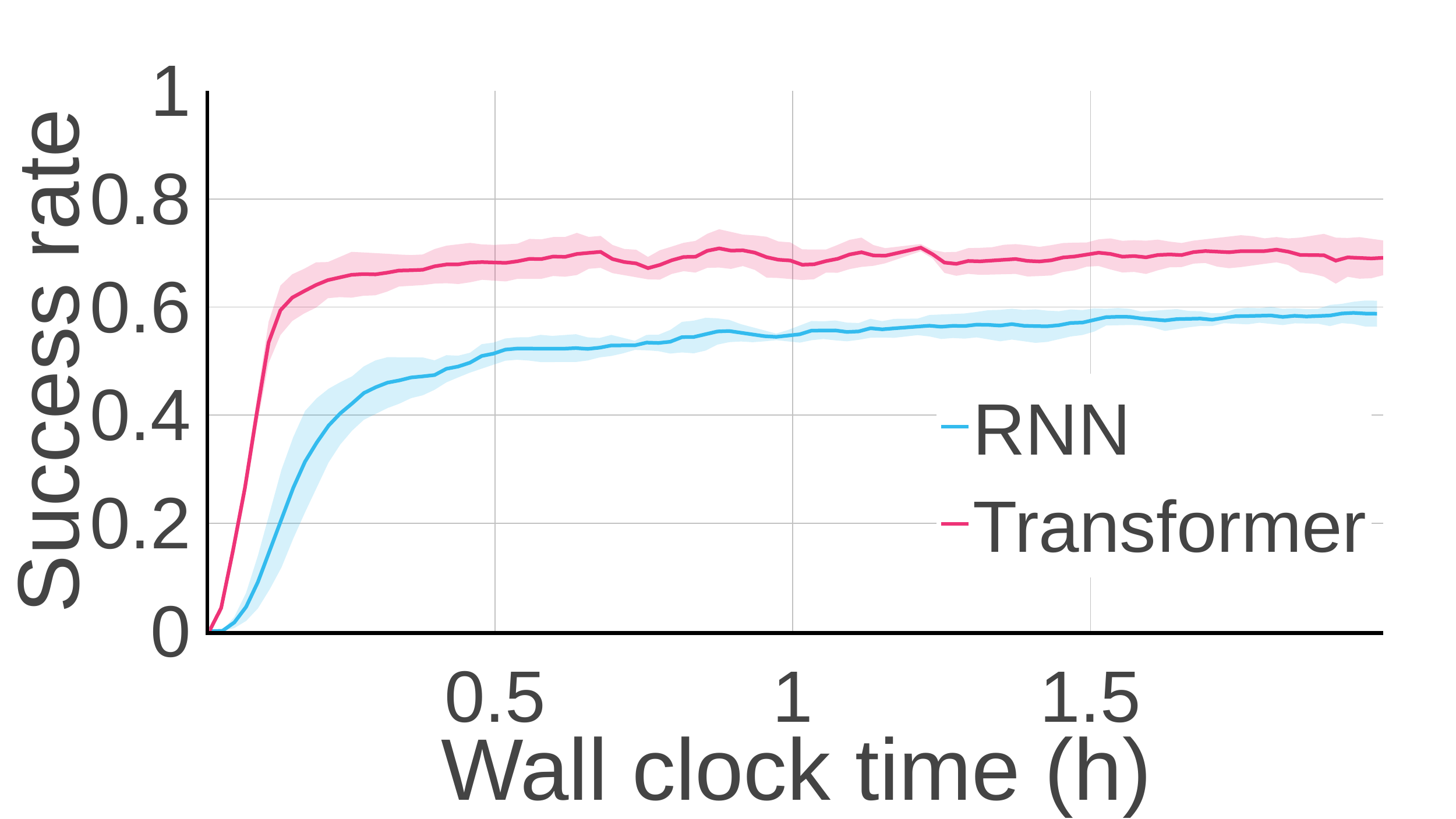}
\caption{}
\label{fig:transformer_time}
\end{subfigure}
\caption{Learning curves of student policies with a Transformer or RNN architecture with respect to the number of samples and wall-clock time, respectively.}
\label{fig:transformer_rnn}
\end{figure}

\paragraph{Transformer vs RNN}
Different from prior works~\cite{andrychowicz2020learning,handa2022dextreme,chen2021system,chen2022visual}, our student policy uses a Transformer architecture instead of an RNN architecture. We compared the learning performance of a Transformer-based policy and an RNN-based policy. \figref{fig:transformer_step} and \figref{fig:transformer_time} show that a Transformer-based policy learns much faster and gets better performance at convergence than an RNN-based policy.

\end{appendices}
\end{document}